\pdfoutput=1

\documentclass[11pt]{article}

\usepackage[final]{acl}

\usepackage{times}
\usepackage{latexsym}
\usepackage{booktabs}
\usepackage{multirow}
\usepackage{adjustbox}
\usepackage{amsmath}
\usepackage{makecell}
\usepackage{float}
\usepackage{amssymb}
\usepackage{listings}
\usepackage{xcolor}
\usepackage{tcolorbox}
\tcbuselibrary{skins, breakable}
\usepackage{multicol} 
\usepackage{subcaption}
\usepackage{pgfplotstable}
\usepackage{pgfplots}
\usepackage{caption}
\pgfplotsset{compat=1.18}
\usepackage{fancyvrb}
\usepackage{enumitem}
\usepackage{hyperref}
\usepgfplotslibrary{groupplots} 
\usepackage[T1]{fontenc}

\usepackage[utf8]{inputenc}

\usepackage{microtype}

\usepackage{inconsolata}
\usepackage{pgfplots}
\usepackage{pgfplotstable}
\usepackage{graphicx}

\usepackage{amsmath}
\usepackage{algorithm}
\usepackage{algpseudocode}
\usepackage[english]{babel}
\title{RTQA : Recursive Thinking for Complex Temporal Knowledge Graph Question Answering with Large Language Models}

\author{
  Zhaoyan Gong\textsuperscript{$\spadesuit$},
  Juan Li\textsuperscript{$\spadesuit$},
  Zhiqiang Liu\textsuperscript{$\spadesuit$$\diamondsuit$},
  Lei Liang\textsuperscript{$\clubsuit$$\diamondsuit$},
  Huajun Chen\textsuperscript{$\spadesuit$$\diamondsuit$},
  Wen Zhang\textsuperscript{$\spadesuit$$\diamondsuit$\textdagger}
  \\
  \textsuperscript{$\spadesuit$} Zhejiang University
  \textsuperscript{$\clubsuit$} Ant Group \\
  \textsuperscript{$\diamondsuit$} ZJU-Ant Group Joint Lab of Knowledge Graph    \\
  \texttt{\{gongzhaoyan,zhang.wen\}@zju.edu.cn}
}

\begin{document}
\maketitle

\begingroup
  \renewcommand{\thefootnote}{\fnsymbol{footnote}}
  \setcounter{footnote}{0}
  \footnotetext{\textsuperscript{\textdagger} Corresponding author}
\endgroup

\begin{abstract}
Current temporal knowledge graph question answering (TKGQA) methods primarily focus on implicit temporal constraints, lacking the capability of handling more complex temporal queries, and struggle with limited reasoning abilities and error propagation in decomposition frameworks. We propose \textbf{RTQA}, a novel framework to address these challenges by enhancing reasoning over TKGs without requiring training. Following recursive thinking, RTQA recursively decomposes questions into sub-problems, solves them bottom-up using LLMs and TKG knowledge, and employs multi-path answer aggregation to improve fault tolerance. RTQA consists of three core components: the Temporal Question Decomposer, the Recursive Solver, and the Answer Aggregator. Experiments on MultiTQ and TimelineKGQA benchmarks demonstrate significant Hits@1 improvements in "\textit{Multiple}" and "\textit{Complex}" categories, outperforming state-of-the-art methods. Our code and data are available at  \href{https://github.com/zjukg/RTQA}{https://github.com/zjukg/RTQA}.

\end{abstract}

\section{Introduction}
In the real world, entities and relationships evolve dynamically, making Temporal Knowledge Graphs (TKGs) with time-aware quadruples more challenging yet practically significant for Question Answering (QA) compared to static Knowledge Graphs (KGs)~\cite{chen2024large}. For instance, the question "\textit{Who is the US President in 2025?}" can be answered using the quadruple \textit{(Trump, president of, United States, 2025)}, representing a simple temporal query.

\begin{figure}[t]
    \centering
    \includegraphics[width=\linewidth]{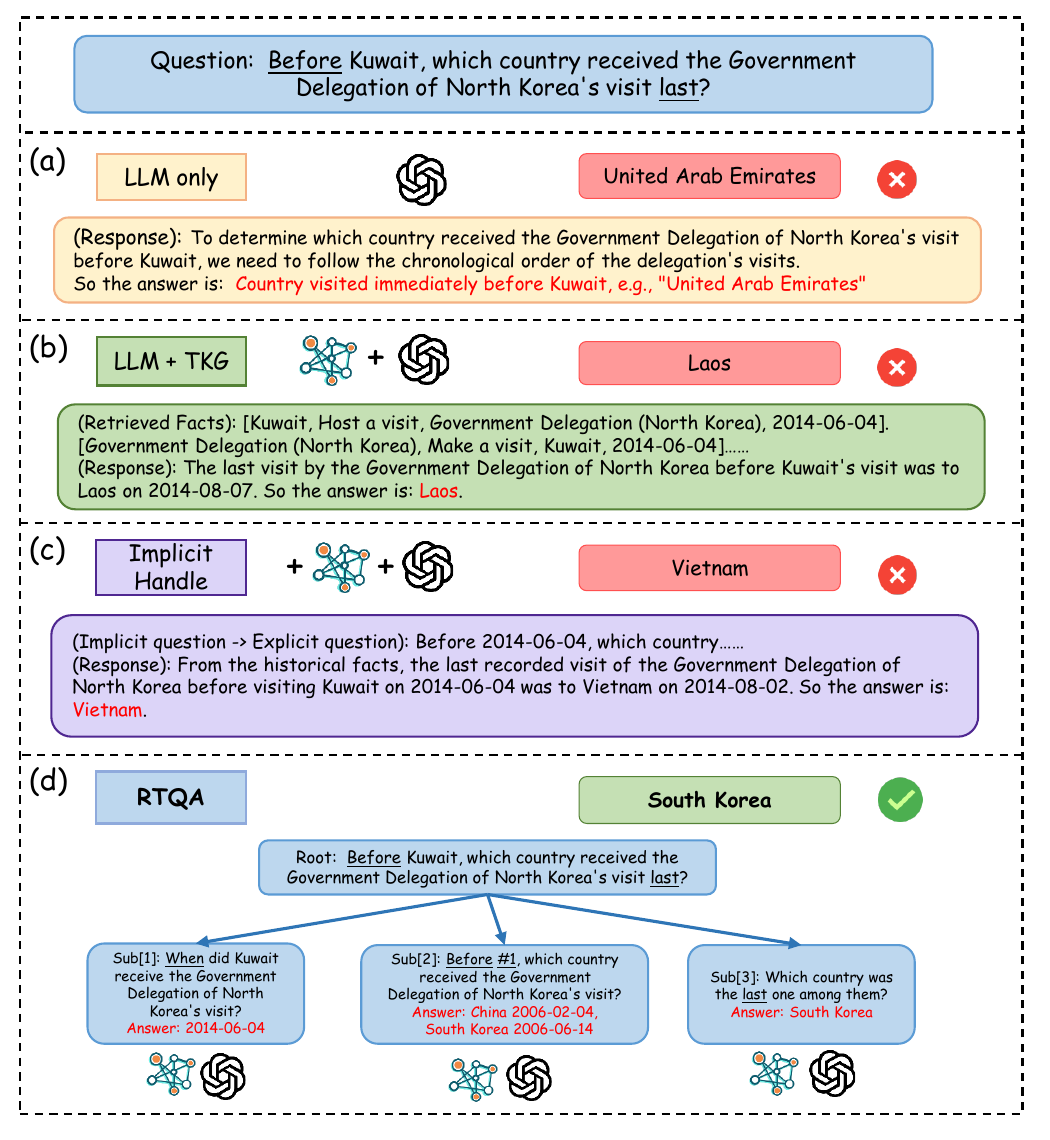}
     \vspace{-2em}
    \caption{
    Motivation comparison: Prior methods (a--c) fail at multi-constraint reasoning, while (d) \textbf{RTQA} solves it via recursive sub-question decomposition.
    }
    \label{fig:intro2}
    \vspace{-1em}
\end{figure}

Recent TKGQA research targets complex queries, including implicit temporal constraints, multi-constraint combinations, multi-hop reasoning, and multi-granular time, as exemplified by the question in Figure~\ref{fig:intro2}: "\textit{Before Kuwait, which country received the Government Delegation of North Korea's visit last?}" Such queries are highly relevant to real-world applications.

While prior works have focused on simple~\cite{saxena2021question,mavromatis2022tempoqr} or implicit temporal questions~\cite{chen2022temporal,qian2024timer4,jia2024faithful}, the integration of Large Language Models (LLMs) into TKGQA offers new opportunities due to their excellent reasoning ability. However, two key challenges persist:

\textbf{(1) Limited reasoning for complex temporal queries.} LLMs often hallucinate~\cite{azaria2024chat,zhao2024generating} when addressing intricate questions. As shown in Figure~\ref{fig:intro2}(a), relying solely on internal knowledge yields incorrect answers like "\textit{United Arab Emirates}". (b) Incorporating TKG facts resolves simpler queries but fails to handle implicit constraints like "\textit{before Kuwait}", producing errors such as "\textit{Laos}". (c) Single-round question rewriting converts implicit constraints to explicit timestamps (e.g., "\textit{before 2014-06-04}"), but struggles with combined constraints like "\textit{before/last}", leading to incorrect answers like "\textit{Vietnam}". Developing frameworks for multi-fact and multi-constraint temporal reasoning remains critical.

\textbf{(2) Error propagation in decomposition frameworks.} Existing methods lack fault tolerance, allowing sub-question errors to propagate. For example, in Figure~\ref{fig:introxr}(b), the query "\textit{When Stalin ended his leadership in his own country, what job did Churchill work for?}" is decomposed into sub-questions: "\textit{Stalin was the leadership of which country?}" yields "\textit{Soviet Union}", followed by "\textit{When was Stalin end his leadership in \#1\footnote{\footnotesize \#1 is a placeholder for the answer ``Soviet Union'' to Sub[1]: ``Which country did Stalin lead?''}?}" answered incorrectly as "\textit{1929}", leading to the erroneous final answer "\textit{Chancellor of the Exchequer}" for "\textit{When \#2\footnote{\footnotesize \#2 is anather placeholder, representing the answer "1929" to the Sub[2]: "When was Stalin end his leadership in \#1?"}, what job did Churchill work for?}". Addressing error propagation issues and building more robust frameworks that enhance fault tolerance in LLM reasoning or fact retrieval processes represents another significant challenge.

To address these challenges, we introduce \textbf{RTQA} (\textbf{R}ecursive \textbf{T}emporal Knowledge Graph \textbf{Q}uestion \textbf{A}nswering), a novel TKGQA framework that decomposes complex temporal questions into sub-questions and performs recursive bottom-up reasoning. By integrating external knowledge from TKGs, RTQA enhances LLMs' ability to tackle intricate temporal queries.

Following a divide-and-conquer strategy, RTQA mimics human problem-solving by breaking down complex questions into manageable parts. As shown in Figure~\ref{fig:intro2}(d), a question is split into three sub-questions: extracting implicit time, applying a ``before'' constraint, and applying a ``last'' constraint. The answer to Sub[1] (``2014-06-04'') informs Sub[2], which generates entity-time pairs (e.g., ``China 2008-02-04, South Korea 2006-06-14'') satisfying the ``before \#1'' constraint. Sub[3] then selects the entity that meets the ``last'' constraint, producing the final answer, ``South Korea''.

To reduce error propagation in sub-questions, we designed a multi-path answer aggregation module that combines answers from both sub-questions and the original question, selecting the most reliable response. As illustrated in Figure~\ref{fig:introxr}(a), we define \texttt{IR\_answer} and \texttt{child\_answer}. Atomic questions (sub[3], sub[4]) rely on a single answer source, while non-atomic questions (sub[1], sub[2], Root) aggregate multiple sources. For instance, when sub[4] incorrectly outputs “\textit{1929},” sub[1]’s \texttt{IR\_answer} correctly identifies “\textit{1953},” preventing error propagation and ensuring the accurate final answer, “\textit{the Prime Minister of the UK}” instead of the wrong “\textit{Chancellor of the Exchequer}.”

We conduct extensive experiments on two challenging TKGQA benchmarks. RTQA consistently outperforms state-of-the-art methods, with notable gains in the "\textit{Multiple}" and "\textit{Complex}" categories. Our main contributions are summarized as follows:
\begin{figure}[t]
    \centering
    \includegraphics[width=1\linewidth]{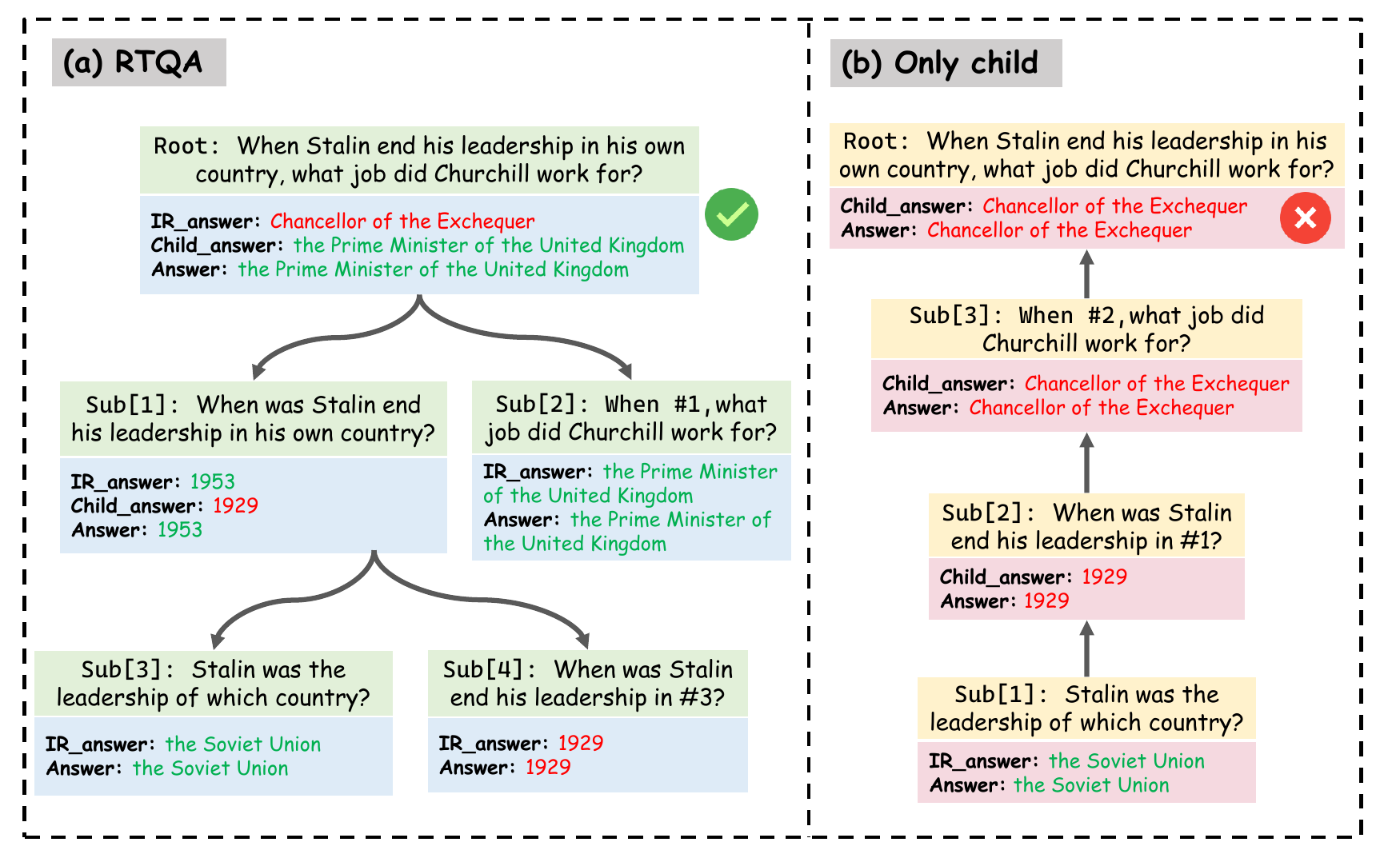}
    \vspace{-1.5em}
    \caption{
    Comparison of \textbf{RTQA} and \textbf{Only-Child} strategies. RTQA mitigates error propagation by integrating child\_answer with IR\_answer, while Only-Child relies solely on child\_answer, compounding earlier errors.
}
    \label{fig:introxr}
    \vspace{-1em}
\end{figure}
\vspace{-1em}
\begin{itemize}
    \item We introduce a framework that recursively decomposes complex temporal questions into sub-questions, reasoning bottom-up to derive accurate answers.
    \vspace{-1em}
    \item We aggregate answers from multiple sources for each question \textit{original, intermediate, atomic}, mitigating error propagation and enhancing framework robustness.
    \vspace{-1em}
    \item Our training-free, plug-and-play approach requires no computational overhead, adapts to various large models, and demonstrates significant performance gains in complex temporal question answering.
\end{itemize}

\section{Related Work}

\subsection{TKGQA}

TKGQA methods can be categorized into semantic parsing-based approaches and embedding-based approaches, with a recent emergence of methods leveraging large language models.

\paragraph{Semantic Parsing-based methods}
Semantic parsing-based methods convert natural language questions into logical expressions to query TKGs, as seen in TEQUILA~\cite{jia2018tequila}, SYGMA~\cite{neelam2021sygma}, SF-TQA~\cite{ding2022semantic}, and Prog-TQA~\cite{chen2024self}. These approaches offer high accuracy when queries are well-formed but struggle with complex questions due to syntax errors in logical expressions, leading to query failures.

\paragraph{Embedding-based Methods}
\begin{figure*}
    \centering
    \includegraphics[width=1\linewidth]{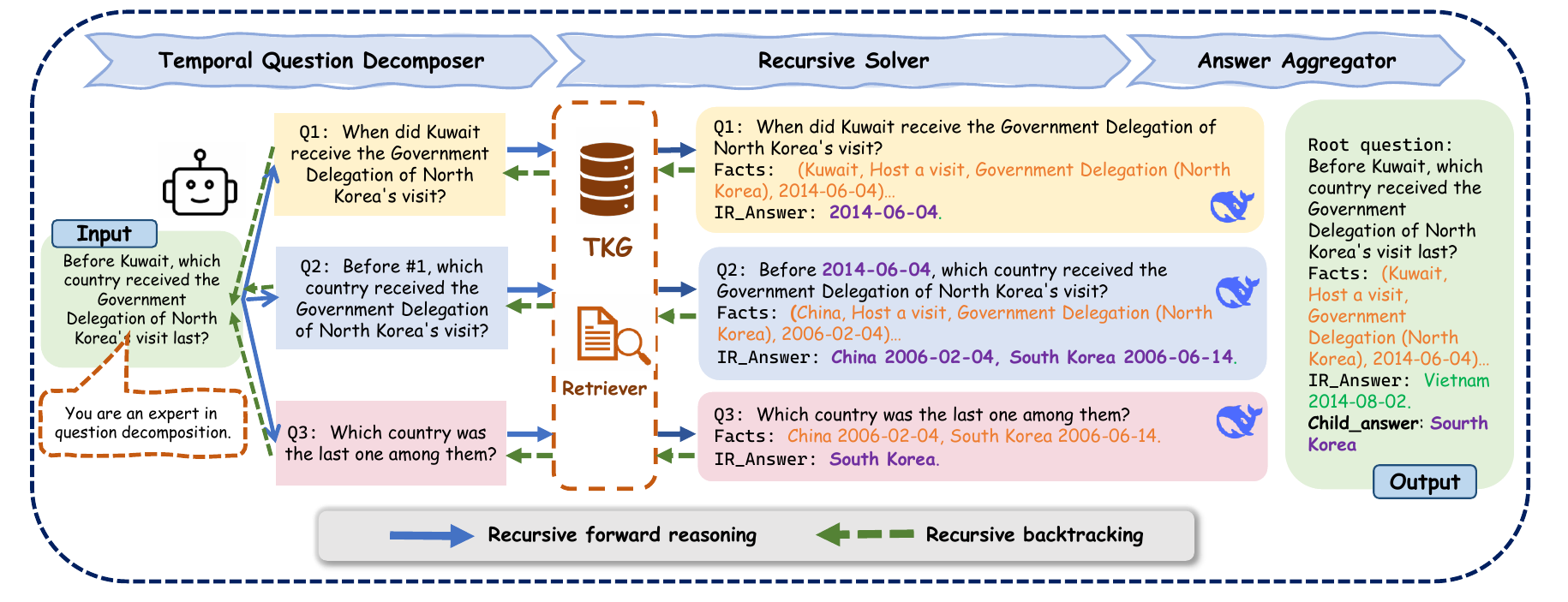}
    \caption{
    An illustration of the \textbf{RTQA} framework applied to a complex temporal question. The framework consists of three stages: (I) \textbf{Temporal Question Decomposer}, which breaks down the original query into sub-questions with explicit temporal constraints; (II) \textbf{Recursive Solver}, where each sub-question is solved using an LLM and retrieved TKG facts; and (III) \textbf{Answer Aggregator}, which integrates the sub-answers to produce the final answer. The reasoning process follows a bottom-up recursive traversal from the root of the decomposition tree, enabling robust aggregation of intermediate results.
}
    \label{FIG:MAIN}
\end{figure*}
TKG Embedding-based methods encode questions and TKG quadruples as low-dimensional vectors, ranking answers by vector semantic similarity. CronKGQA~\cite{saxena2021question} introduces learnable reasoning, TempoQR~\cite{mavromatis2022tempoqr} enhances embeddings with contextual and temporal modules, and MultiQA~\cite{chen2023multi} aggregates multi-granular time information. Other approaches incorporate graph neural networks~\cite{jia2024faithful,liu2023local,sharma2022twirgcn}. These methods ensure high execution rates but can only handle simple questions and perform poorly on complex temporal questions.

\paragraph{LLM-based Methods}
Recent LLM-based approaches, such as ARI~\cite{chen2023temporal}, GenTKGQA~\cite{gao2024two}, FAITH~\cite{jia2024faithful}, and TimeR4~\cite{qian2024timer4}, leverage LLMs for TKGQA. Unlike these methods, which often require retraining, our \textbf{RTQA} framework is training-free and plug-and-play, handling complex queries with multiple entities, multi-hop reasoning, and compound temporal constraints while maintaining compatibility with various LLMs.

\subsection{Question decomposing}

Question decomposition emulates human problem-solving by breaking complex queries into simpler sub-questions, a strategy effective for multi-hop reasoning in KGQA~\cite{cao2020kqa,khot2022decomposed,trivedi2022musique,cao-etal-2023-probabilistic}. However, existing approaches inadequately address temporal questions, necessitating advanced frameworks like RTQA.

\section{Preliminary}

\textbf{Temporal constraint} defines a condition related to a specific time point or interval that must be met by both the answer and its supporting evidence. This includes 13 Allen temporal relations~\cite{allen1984towards}, 3 temporal set relations, duration comparisons, and sorting mechanisms~\cite{sun2025timelinekgqa}.

\textbf{TKG} A temporal knowledge graph \(\mathcal{G} = \{\mathcal{E}, \mathcal{P}, \mathcal{T}, \mathcal{F}\}\) is a directed graph where vertices are a set of entities \(\mathcal{E}\). The edges are a set of predicates \(\mathcal{P}\) with timestamps \(\mathcal{T}\). The quadruple set 
\(\mathcal{F} = \{(s, p, o, t) \mid \mathcal{E} \times \mathcal{P} \times \mathcal{E} \times \mathcal{T}\}\) 
represents the temporal facts, where \(s\) and \(o\) are subject and object, respectively, and \(p\) is the predicate between \(s\) and \(o\) at timestamp \(t\).

\textbf{TKGQA} is a task to infer the correct answer to a natural language question \(q \in \mathcal{Q}\) based on relevant quadruples \(f = (s, p, o, t)\) in the TKG, where the answer can be either an entity name or a timestamp.

\section{Method}

\subsection{Method Overview}
Inspired by the divide-and-conquer principle, RTQA enables efficient handling of complex temporal dependencies. As shown in Figure~\ref{FIG:MAIN}, \textbf{Temporal Question Decomposer} (Section~\ref{section:question decom}) firstly transforms complex temporal questions into a series of simpler sub-questions by identifying implicit temporal constraints (e.g., "\textit{before}", "\textit{last}") and converting them into explicit temporal expressions. It also extracts relevant multi-hop facts and temporal granularity information. For example, the question \textit{``Before Kuwait, which country received the Government Delegation of North Korea's visit last?''} is decomposed into three sub-questions: (1) \textit{When did Kuwait receive the visit?} (2) \textit{Which countries received the visit before \#1?} (3) \textit{Which was the latest among them?} 
Next, \textbf{Recursive Solver} (Section~\ref{section:recursive}) leverages the reasoning ability of LLMs and the factual knowledge from the TKG to recursively solve sub-questions in a bottom-up manner. The resulting answers are used to replace placeholders in parent questions, forming a progressive reasoning chain. For instance, the answer \textit{``2014-06-04''} to sub-question (1) serves as the temporal reference for sub-question (2), which then filters out earlier visits, and sub-question (3) selects the latest one from the filtered list. This recursive approach effectively handles both implicit and compound temporal constraints.
Finally, \textbf{Answer Aggregator} (Section~\ref{section:answer}) consolidates results, evaluating candidates (e.g., "\textit{South Korea}" vs. "\textit{Vietnam}") to ensure accuracy and robust fault tolerance.

\subsection{Temporal Question Decomposer}\label{section:question decom}
The goal of this stage is to decompose a complex temporal question \( Q \) into a series of sub-questions \( T \), where \( Q \) is the root node in \( T \). Basically, as shown in Figure~\ref{FIG:MAIN}, the query tree is generated by LLMs with few-shot prompting. 

\textbf{The question decomposition process} can be formalized as a sequence of transformations applied to an input question \( Q \). Let the instruction template be denoted as \( \mathcal{I} \), and the question type as \( \tau = \textbf{Type}(Q) \), where \( \text{Type}(\cdot) \) is the type identification function. The prompt is constructed and the LLM response is obtained as follows:
\begin{align}
    p &\leftarrow \text{BuildPrompt}(Q, \tau, \mathcal{I}),\\
    r_{\text{llm}} &\leftarrow \text{LLMCaller}(Q,p),
\end{align}
where \( \text{BuildPrompt}(\cdot) \) denotes the prompt construction function, and \( \text{LLMCaller}(\cdot) \) represents the LLM call function. The structured response is then parsed, and the temporal decomposition tree is constructed:
\begin{align}
    S &\leftarrow \text{ParseStruct}(r_{\text{llm}}), \\
    \mathcal{T} &\leftarrow \text{BuildTree}(S),
\end{align}
where \( \text{ParseStruct}(\cdot) \) extracts structured elements from the LLM output, and \( \text{BuildTree}(\cdot) \) organizes them into a hierarchical decomposition tree \( \mathcal{T} \).

\textbf{Each node \( q^i \in \mathcal{T} \) contains:} (i) a node index \( \texttt{idx} \), (ii) the question text \( \texttt{question\_text} \), (iii) a list of child nodes \( \texttt{sons} \), (iv) the parent node index \( \texttt{fa} \), (v) metadata including the question type label \( \texttt{qlabel} \), and (vi) the \( \texttt{gold\_answer} \). This structure maintains hierarchical and semantic fidelity throughout the reasoning process.

\textbf{The construction of the prompts} is tailored to various types of temporal questions. For each type, 5--10 question examples are carefully selected from the validation set, with their sub-question decompositions manually crafted. The specific prompts constructed for each category, along with their corresponding decompositions, are illustrated in the Figure~\ref{fig:prompt equal},~\ref{fig:prompt before after}, provided in the Appendix~\ref{A:prompts for decom} for more details.

\subsection{Recursive Solver}\label{section:recursive}

\textbf{Recursive solving process.}
We adopt a recursive post-order traversal to solve the query decomposition tree, starting from the root and proceeding in a bottom-up manner. The solver is formalized as a unified recursive function \( \text{Solve}(q^i, \mathcal{T}, \mathcal{R}, \theta) \), where \( \mathcal{R} \) denotes the retriever for TKG grounding, and \( \theta \) denotes the reasoning LLM.

For a leaf node \( q^i \in \mathcal{T} \), the solver first retrieves relevant facts from the TKG and then invokes the LLM to generate an answer, as defined below:
\begin{align}
\mathcal{F}^i &\leftarrow \text{Retrieve}(q^i, \mathcal{R}), \\
a^i &\leftarrow \text{Reason}(q^i, \mathcal{F}^i, \theta).
\end{align}

For a non-leaf node \( q^i \), the solver recursively processes each child \( q^{c_j} \in \texttt{sons}(q^i) \), where \( j = 1,\dots,n \) and \( n \) is the number of sub-questions. For the first child, the answer is computed directly:
\begin{align}
q^{c_1}_{\text{updated}} &\leftarrow q^{c_1}, \\
a^{c_1} &\leftarrow \text{Solve}(q^{c_1}_{\text{updated}}, \mathcal{T}, \mathcal{R}, \theta).
\end{align}
Subsequent questions are updated by replacing placeholders (e.g., \(\#k\)) with prior answer \(a^k\):
\begin{align}
q^{c_2}_{\text{updated}} &\leftarrow \text{Replace}(q^{c_2}, \{a^{c_1}\}), \\
a^{c_2} &\leftarrow \text{Solve}(q^{c_2}_{\text{updated}}, \mathcal{T}, \mathcal{R}, \theta), \\
&\vdots \\
q^{c_n}_{\text{updated}} &\leftarrow \text{Replace}(q^{c_n}, \{a^{c_1}, \ldots, a^{c_{n-1}}\}), \\
a^{c_n} &\leftarrow \text{Solve}(q^{c_n}_{\text{updated}}, \mathcal{T}, \mathcal{R}, \theta),
\end{align}
where \(\{a^{c_1}, \ldots, a^{c_{j-1}}\}\) denotes answers from prior sub-questions used for reference replacement.

After solving all sub-questions, the final answer of this non-leaf node \( q^i \) is aggregated via a summary function:
\begin{equation}
a_{child}^i = \text{Summarize}(q^i, \{a^{c_1}, \ldots, a^{c_n}\}).
\end{equation}
This recursive procedure ensures consistent resolution of complex temporal queries across all levels of the tree.

\textbf{Relevant Facts Retriever.} 
The quadruples or quintuples in TKG are converted into natural language statements in the following two forms:
\begin{center}
\begin{tcolorbox}[colback=gray!10, boxrule=0pt, arc=1pt, width=0.48\textwidth, height=1cm, halign=left, valign=center]
\begin{Verbatim}[fontsize=\scriptsize]
{subject} {predicate} {object} in {time}
{subject} {predicate} {object} from {start} to {end}
\end{Verbatim}
\end{tcolorbox}
\end{center}
The statements are then embedded using a dense encoder along
with input question. The top-$K$ most relevant facts are retrieved based on similarity.

\textbf{Explainable reasoning with LLM.}
The RTQA framework employs a post-processing module to distill concise, standardized answers from LLM outputs. Step-by-step reasoning, guided by precise instructions, ensures transparency by preserving the full inference chain, a hallmark of RTQA’s interpretability. The LLM concludes with a structured summary, \textit{So the answer is:}, enabling reliable extraction of the final entity or timestamp while retaining the reasoning for clarity. The prompts driving this process are detailed in Appendix~\ref{A:prompts for solver} Figure~\ref{fig:prompt solver historal},~\ref{fig:prompt solver relevant}.

\textbf{Time Expression Standardization.}
Before invoking the recursive solver, all time expressions are standardized to the ISO 8601\footnote{\footnotesize \url{https://www.iso.org/iso-8601-date-and-time-format.html}} format (\texttt{yyyy-mm-dd}). This preprocessing step ensures consistent handling of temporal references across different granularities (year, month, day), addressing the variability in natural language expressions and improving the accuracy of temporal reasoning.

\subsection{Answer Aggregator}\label{section:answer}

The aggregator selects the most plausible final answer by fusing two candidates:  \( a^i_{\text{IR}} \) and \( a^i_{\text{child}} \). Specifically, \( a^i_{\text{IR}} \) is produced by retrieving relevant TKG facts and applying LLM reasoning. \( a^i_{\text{child}} \) aggregates answers from the child nodes of the query tree. 
In cases of ambiguity, the aggregator leverages the original query context to choose the most appropriate answer, ensuring alignment with user intent. The detailed prompt design for this aggregation process is provided in Appendix~\ref{A:prompts for answer aggre} Figure~\ref{fig:prompt answer aggre}. This process is formalized as:
\begin{equation}
    a_{\text{final}}^i = \text{Aggregator}(a^i_{\text{IR}}, a^i_{\text{child}}).
\end{equation}
In conclusion, the answer aggregator serves as a critical module to prevent errors from propagating upstream by selecting one of three answer sources as the final answer.

\section{Experiment}
\begin{table*}
\caption{Performance comparison of baselines and RTQA on Hits@1 and Hits@10 across various question types and answer types on MultiTQ testset. The best and second best results are marked in \textbf{bold} and \underline{underlined}, respectively.}
\label{tab:mainresult_multitq}
\vspace{-0.5em}
\centering
\small
\setlength{\tabcolsep}{5pt}  
\begin{adjustbox}{max width=\textwidth}
\begin{tabular}{lcccccccccccc}
\toprule
\multirow{3}{2cm}{\textbf{Model}} 
& \multicolumn{5}{c}{\textbf{Hits@1}} 
& \multicolumn{5}{c}{\textbf{Hits@10}} \\
\cmidrule(lr){2-6} \cmidrule(lr){7-11}
& \textbf{Overall} 
& \multicolumn{2}{c}{\textbf{Question Type}} 
& \multicolumn{2}{c}{\textbf{Answer Type}} 
& \textbf{Overall} 
& \multicolumn{2}{c}{\textbf{Question Type}} 
& \multicolumn{2}{c}{\textbf{Answer Type}} \\
\cmidrule(lr){3-4} \cmidrule(lr){5-6}
\cmidrule(lr){8-9} \cmidrule(lr){10-11}
& & \textbf{Multiple} & \textbf{Single} & \textbf{Entity} & \textbf{Time} 
& & \textbf{Multiple} & \textbf{Single} & \textbf{Entity} & \textbf{Time} \\
\midrule
BERT         & 0.083 & 0.061 & 0.092 & 0.101 & 0.040 & 0.441 & 0.392 & 0.461 & 0.531 & 0.222 \\
DistillBERT  & 0.083 & 0.074 & 0.087 & 0.102 & 0.037 & 0.482 & 0.426 & 0.505 & 0.591 & 0.216 \\
ALBERT       & 0.108 & 0.086 & 0.116 & 0.139 & 0.032 & 0.484 & 0.415 & 0.512 & 0.589 & 0.228 \\
LLaMA2       & 0.185 & 0.101 & 0.220 & 0.239 & 0.055 & - & - & - & - & - \\
ChatGPT      & 0.102 & 0.077 & 0.147 & 0.137 & 0.020 & - & - & - & - & - \\
\midrule
EmbedKGQA    & 0.206 & 0.134 & 0.235 & 0.290 & 0.001 & 0.459 & 0.439 & 0.467 & 0.648 & 0.001 \\
CronKGQA     & 0.279 & 0.134 & 0.337 & 0.328 & 0.156 & 0.608 & \underline{0.453} & 0.671 & 0.696 & 0.392 \\
MultiQA     & 0.293 & 0.159 & 0.347 & 0.349 & 0.157 & \underline{0.635} & \textbf{0.519} & \underline{0.682} & \textbf{0.733} & \underline{0.396} \\
\midrule
ARI     & 0.380 & 0.210 & 0.680 & 0.394 & 0.344 & - & - & - & - & - \\
TimeR4     & \underline{0.728} & \underline{0.335} & \underline{0.887} & \underline{0.639} & \textbf{0.945} & - & - & - & - & - \\
\midrule
\textbf{RTQA} & \textbf{0.765} & \textbf{0.424} & \textbf{0.902} & \textbf{0.692} & \underline{0.942} 
                & \textbf{0.768}& 0.427 & \textbf{0.907} & \underline{0.697} & \textbf{0.942} \\
\bottomrule
\end{tabular}
\end{adjustbox}

\vspace{-0.5em}
\end{table*}

\begin{table}[t] 
\caption{Performance(Hits@1) comparison of RAG baseline and RTQA across \textit{Simple}, \textit{Medium}, \textit{Complex} on TimelineKGQA test dataset.}
\label{tab:mainresult_TimelineKGQA}
\vspace{-0.5em}
\centering
\small
\setlength{\tabcolsep}{1pt}
\begin{adjustbox}{max width=\textwidth}
\begin{tabular}{c cccc}
\toprule
\multirow{2}{*}{\textbf{Model}} & \multicolumn{4}{c}{\textbf{Hits@1}} \\
\cmidrule(lr){2-5} 
& \textbf{Overall} & \textbf{Simple} & \textbf{Medium} & \textbf{Complex} \\
\midrule
RAG baseline & 0.235 & 0.704 & 0.092 & 0.009 \\
\textbf{RTQA} & \textbf{0.298} & 0.608 & \textbf{0.218} & \textbf{0.135}  \\
\bottomrule
\end{tabular}
\end{adjustbox}

\vspace{-0.5em}
\end{table}

\subsection{Experimental Setup}
\paragraph{Datasets}
We evaluate RTQA on two challenging TKGQA benchmarks: \textsc{MultiTQ}~\cite{chen2023multi} and \textsc{TimelineKGQA}~\cite{sun2025timelinekgqa}. \textsc{MultiTQ} offers large-scale QA pairs with diverse temporal granularities, while \textsc{TimelineKGQA} covers questions with varying complexity and time formats. The test sets contain 54,584 and 8,344 questions, respectively. Detailed statistics and category distributions are provided in Appendix~\ref{A:dataset}.

\paragraph{Baselines}
We compare RTQA against three types of baselines on \textsc{MultiTQ}: (1) \textbf{Pre-trained LMs}, including BERT~\cite{devlin2019bert}, DistillBERT~\cite{sanh2019distilbert}, ALBERT~\cite{lan2019albert}, LLaMA2~\cite{touvron2023llama}, and ChatGPT; (2) \textbf{TKG embedding-based methods}, including EmbedKGQA~\cite{saxena2020improving}, CronKGQA~\cite{saxena2021question}, and MultiQA~\cite{chen2023multi}; (3) \textbf{LLM-based methods}, including ARI~\cite{chen2023temporal} and TimeR4~\cite{qian2024timer4}.
For \textsc{TimelineKGQA}, due to its complexity, existing embedding-based models are not directly applicable. Following~\cite{sun2025timelinekgqa}, we adopt a Retrieval-Augmented Generation (RAG) baseline.

\paragraph{Implementation Details}
We used the \textsc{OpenAI} API (gpt-4o-mini\footnote{\footnotesize \url{https://platform.openai.com/docs/models/gpt-4o-mini}}) for temporal question decomposition, and the \textsc{Deepseek} API (deepseek-v3-250324) for answer reasoning on the MultiTQ dataset. For TimelineKGQA, all stages used \textsc{OpenAI} (gpt-4o-mini). The temperature was set to 0 for deterministic outputs. We employed the BGE-M3\footnote{\footnotesize \url{https://huggingface.co/BAAI/bge-m3}}~\cite{chen2024bge} model via Hugging Face to generate dense embeddings of TKG triples and questions, though hybrid retrieval was not used. Dense retrieval and clustering were performed using \textsc{FAISS}~\cite{douze2024faiss}, following~\cite{qian2024timer4}. To avoid excessive context, we limited reasoning inputs to the top 50 retrieved facts.

\subsection{Main Results}
We present the experimental results in comparisons between our model and existing state-of-the-art baseline models on the MultiTQ and TimelineKGQA datasets in Table~\ref{tab:mainresult_multitq} and Table~\ref{tab:mainresult_TimelineKGQA}.

\paragraph{Performance Comparison on MultiTQ}  
We evaluate performance using Hits@1 and Hits@10, with breakdowns by question type (multiple, single) and answer type (entity, time).\footnote{\footnotesize Baseline results are from~\cite{qian2024timer4}.}  
As shown in Table~\ref{tab:mainresult_multitq}, RTQA outperforms all baselines across nearly all metrics. It achieves a Hits@1 of 0.765, surpassing the second-best model TimeR4 (0.728). For question types, RTQA scores 0.424 on multiple and 0.902 on single, demonstrating strong adaptability to varying complexities. On Hits@10, RTQA maintains the lead with 0.768 overall, and excels on time answers with a score of 0.942, highlighting its effectiveness in handling temporal reasoning in TKGQA.
\begin{table}[t]
\caption{Ablation studies of RTQA on MultiTQ.}
\label{tab:ablation result}
\vspace{-0.5em}
\centering
\footnotesize
\setlength{\tabcolsep}{2pt}
\begin{adjustbox}{max width=0.8\textwidth} 
\begin{tabular}{lccccc}
\toprule
\multirow{2}{1cm}{\textbf{Model}} & \multirow{2}{*}{\textbf{Overall}} & \multicolumn{2}{c}{\textbf{Question Type}} & \multicolumn{2}{c}{\textbf{Answer Type}} \\
\cmidrule(lr){3-4} \cmidrule(lr){5-6}
& & \textbf{Multiple} & \textbf{Single} & \textbf{Entity} & \textbf{Time} \\
\midrule
\textbf{RTQA} & \textbf{0.765} & \textbf{0.424} & \textbf{0.902} & \textbf{0.692} & \textbf{0.942} \\
w/o decomposer & 0.709 & 0.214 & 0.890 & 0.596 & 0.958 \\
w/o multi-answer  & 0.752 & 0.341 & 0.904 & 0.667 & 0.942 \\
w/o fact retrieval & 0.070 & 0.015 & 0.090 & 0.096 & 0.013 \\
\bottomrule
\end{tabular}
\end{adjustbox}

\vspace{-0.5em}
\end{table}

\begin{table*}[t]
\caption{Experiment results of multi-granular time on Hits@1.}
\label{tab:multi_granularity time}
\vspace{-0.5em}
\centering
\small
\setlength{\tabcolsep}{8pt} 
\begin{adjustbox}{max width=\textwidth} 
\begin{tabular}{lccccccccc}
\toprule
\multirow{2}{*}{\textbf{Model}} & \multicolumn{3}{c}{\textbf{Equal}} & \multicolumn{3}{c}{\textbf{Before/After}} & \multicolumn{3}{c}{\textbf{Equal Multi}} \\
\cmidrule(lr){2-4} \cmidrule(lr){5-7} \cmidrule(lr){8-10}
& \textbf{Day} & \textbf{Month} & \textbf{Year} & \textbf{Day} & \textbf{Month} & \textbf{Year} & \textbf{Day} & \textbf{Month} & \textbf{Year} \\
\midrule
BERT & 0.049 & 0.103 & 0.136 & 0.150 & 0.164 & 0.175 & 0.064 & 0.102 & 0.090 \\
DistillBERT & 0.041 & 0.087 & 0.113 & 0.160 & 0.150 & 0.186 & 0.096 & 0.127 & 0.089 \\
ALBERT & 0.069 & 0.082 & 0.132 & 0.221 & 0.277 & 0.308 & 0.103 & 0.144 & 0.144 \\
EmbedKGQA & 0.200 & 0.336 & 0.218 & \underline{0.392} & 0.518 & 0.511 & 0.145 & 0.321 & 0.263 \\
CronKGQA & 0.425 & 0.389 & 0.331 & 0.375 & 0.474 & 0.450 & 0.295 & \underline{0.333} & 0.251 \\
MultiQA & \underline{0.445} & \underline{0.393} & \underline{0.350} & 0.379 & \underline{0.548} & \underline{0.525} & \underline{0.308} & 0.321 & \underline{0.283} \\
\midrule
\textbf{RTQA} & \textbf{0.916} & \textbf{0.959} & \textbf{0.967} & \textbf{0.842} & \textbf{0.898} & \textbf{0.787} & \textbf{0.729} & \textbf{0.758} & \textbf{0.578} \\
\bottomrule
\end{tabular}
\end{adjustbox}
\vspace{-0.5em}

\end{table*}

Pre-trained language models such as BERT, DistillBERT, and ALBERT perform poorly, with Hits@1 below 0.11, indicating that generic pre-trained models are insufficient for temporal reasoning. While models like EmbedKGQA, CronKGQA, and MultiQA perform reasonably on single-choice and entity questions, they struggle with multiple and time answers. TimeR4, which integrates LLMs for TKGQA, shows better performance but still falls short of RTQA.

\paragraph{Performance Comparison on TimelineKGQA}  
To evaluate the generalization of RTQA, we compare it with the RAG baseline on the TimelineKGQA dataset, focusing on questions of varying complexity: Simple, Medium, and Complex (see Table~\ref{tab:mainresult_TimelineKGQA}).
RTQA achieves an overall Hits@1 of 0.298, outperforming RAG (0.235) by 27\%. Its advantage becomes more pronounced as question complexity increases. On medium-complexity questions, RTQA scores 0.218 v.s.\ RAG's 0.092 (137\% improvement); for complex questions, it reaches 0.135 vs.\ RAG's 0.009, marking a 1400\% gain. 
These results highlight RTQA’s strong capability in complex temporal reasoning, especially on multi-hop questions and those involving intricate time constraints, where traditional methods struggle.

\begin{figure}[t]
\centering
\definecolor{Cteal}{HTML}{E1703C}
\definecolor{Corange}{HTML}{4091CF}
\definecolor{Cpurple}{HTML}{F6C957}
\definecolor{Cgreen}{HTML}{329845}

\begin{tikzpicture}
\begin{groupplot}[
    group style={
        group size=2 by 2,
        horizontal sep=0.7cm,
        vertical sep=0.3cm
    },
    ybar,
    ymin=0, ymax=1.05,
    enlarge x limits=2.0,
    tick style={draw=none},
    xtick=\empty,             
    height=3cm, width=4.5cm,
    ylabel style={font=\small},
    tick label style={font=\tiny},
    title style={font=\small, yshift=-19pt},
    axis line style={line width=0pt}
]

\nextgroupplot[
    ylabel={Hit@1},
    title={Multiple},
    ymin=0.15, ymax=0.55
]
\addplot[ybar, bar width=12pt, fill=Cteal!80,  draw=none] coordinates {(0.2,0.23)};
\addplot[ybar, bar width=12pt, fill=Corange!80,draw=none] coordinates {(0.4,0.38)};
\addplot[ybar, bar width=12pt, fill=Cpurple!80,draw=none] coordinates {(0.6,0.34)};
\addplot[ybar, bar width=12pt, fill=Cgreen!80, draw=none] coordinates {(0.8,0.46)};

\nextgroupplot[
    title={Single},
    ymin=0.80, ymax=0.95
]
\addplot[ybar, bar width=12pt, fill=Cteal!80,  draw=none] coordinates {(0.2,0.85)};
\addplot[ybar, bar width=12pt, fill=Corange!80,draw=none] coordinates {(0.4,0.90)};
\addplot[ybar, bar width=12pt, fill=Cpurple!80,draw=none] coordinates {(0.6,0.92)};
\addplot[ybar, bar width=12pt, fill=Cgreen!80, draw=none] coordinates {(0.8,0.89)};

\nextgroupplot[
    ylabel={Hit@1},
    title={Entity},
    ymin=0.55, ymax=0.75
]
\addplot[ybar, bar width=12pt, fill=Cteal!80,  draw=none] coordinates {(0.2,0.60)};
\addplot[ybar, bar width=12pt, fill=Corange!80,draw=none] coordinates {(0.4,0.67)};
\addplot[ybar, bar width=12pt, fill=Cpurple!80,draw=none] coordinates {(0.6,0.68)};
\addplot[ybar, bar width=12pt, fill=Cgreen!80, draw=none] coordinates {(0.8,0.69)};

\nextgroupplot[
    title={Time},
    ymin=0.85, ymax=1.00,
    title style={yshift=1.5pt}
]
\addplot[ybar, bar width=12pt, fill=Cteal!80,  draw=none] coordinates {(0.2,0.87)};
\addplot[ybar, bar width=12pt, fill=Corange!80,draw=none] coordinates {(0.4,0.96)};
\addplot[ybar, bar width=12pt, fill=Cpurple!80,draw=none] coordinates {(0.6,0.95)};
\addplot[ybar, bar width=12pt, fill=Cgreen!80, draw=none] coordinates {(0.8,0.96)};

\end{groupplot}

\node at ($(current bounding box.south) + (0,-0.3cm)$) {%
\begin{tikzpicture}
  \draw[line width=3pt, Cteal]   (2.5cm,0) -- (2.9cm,0);
  \node[anchor=west, font=\scriptsize] at (2.9cm,0) {gpt-4o-mini};

  \draw[line width=3pt, Corange] (4.6cm,0) -- (5.0cm,0);
  \node[anchor=west, font=\scriptsize] at (5.0cm,0) {gpt-4o};

  \draw[line width=3pt, Cpurple] (6.0cm,0) -- (6.4cm,0);
  \node[anchor=west, font=\scriptsize] at (6.4cm,0) {deepsk-v3};

  \draw[line width=3pt, Cgreen]  (7.7cm,0) -- (8.1cm,0);
  \node[anchor=west, font=\scriptsize] at (8.1cm,0) {deepsk-r1};
\end{tikzpicture}
};

\end{tikzpicture}
\vspace{-0.5em}
\caption{Hits@1 results with different LLMs.}
\label{fig:plug}
\vspace{-0.5em}
\end{figure}
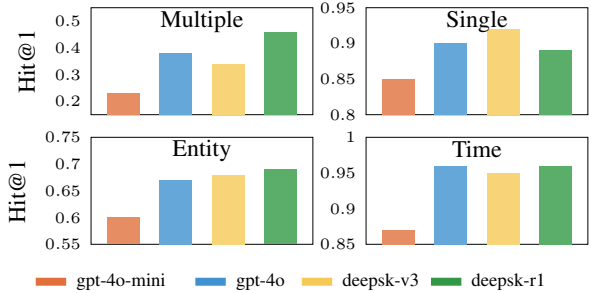

\subsection{Ablation Studies}
To validate the effectiveness of different components in our proposed RTQA model, we conducted a series of ablation studies on the MultiTQ dataset. Table \ref{tab:ablation result} presents the performance of various ablated versions of RTQA, where "w/o" indicates the removal of specific modules.

\paragraph{Impact of Question Decomposition} 
We remove the temporal question decomposer module, processing questions directly without decomposition. As shown in Table \ref{tab:ablation result}, the result drops significantly, with the overall Hits@1 decreasing from $0.765$ to $0.709$.
The impact is particularly pronounced for \texttt{Multiple} questions, where performance drops dramatically from $0.424$ to $0.214$ (a $49.5$\% reduction). This substantial decrease confirms that recursive decomposition is crucial for handling complex temporal reasoning that involves multiple hops or combined temporal constraints.

\paragraph{Impact of Multi-Answer Strategy}

We eliminated the answer aggregator module and evaluated the variant \textit{w/o multi-answer}, which relies solely on answers derived from sub-questions without incorporating alternative sources.
The results show an overall performance drop of 1.7\%, with more pronounced declines for \texttt{Multiple} questions, which decreased by 19.6\%, and \texttt{Entity} answers, which dropped by 3.6\%. These findings highlight the effectiveness of the multi-answer module in reducing error propagation by offering alternative reasoning paths when sub-question inference fails or yields inaccurate results.
\begin{table}[t]
\caption{Characteristics of \(\mathcal{T}\) and API efficiency.}
\label{tab:efficiency}
\centering
\small
\begin{tabular}{lccc}
\toprule
 & \textbf{MultiTQ} & \textbf{TimelineKGQA} \\
\midrule
Avg Depth & 1.37 & 1.57  \\
Avg Branch & 1.60 & 1.81  \\
Avg API Call & 3.96 & 5.38  \\
\bottomrule
\end{tabular}
\vspace{-0.5em}
\end{table}

\paragraph{Impact of Fact Retrieval}

We examined the variant "w/o fact retrieval" that removes external knowledge from TKG, The results reveal a catastrophic performance degradation, with overall Hits@1 plummeting from 0.765 to a mere 0.070.
The magnitude of this performance collapse underscores the fundamental importance of accurate fact retrieval in TKGQA. Without access to reliable factual information, even the most sophisticated reasoning frameworks cannot produce accurate answers, as they lack the necessary evidence base for their inferences.

\subsection{Further Experimental Analysis}
\paragraph{Multi-Granular Time Analysis}
To verify the effectiveness of the model on multi-granularity temporal reasoning, we compared RTQA's performance across different time granularities (day, month, year) and temporal question type (Equal, Before/After, Equal Multi).\footnote{\footnotesize Baseline results are sourced from~\cite{chen2023multi}.} 
Table~\ref{tab:multi_granularity time} demonstrates that RTQA consistently outperforms all baseline models across all temporal granularities and reasoning types. The consistent superior performance demonstrates that our recursive question decomposition approach and multi-answer strategy work effectively regardless of temporal scale, making RTQA a robust solution for diverse temporal reasoning applications.

\paragraph{Generalizability across different LLMs}
To evaluate the adaptability of RTQA across different LLMs, we conducted experiments using \texttt{gpt-4o-mini}, \texttt{gpt-4o}, \texttt{deepseek-v3}, and \texttt{deepseek-r1}. Given the large size of the test set, we randomly sampled 1,000 questions for this study. To ensure consistent input across models, we fixed the question decomposition outputs by using \texttt{gpt-4o-mini} for all decomposition steps, and applied different LLMs only in the \textit{Recursive Solver} stage. 
As shown in Figure~\ref{fig:plug}, models with stronger inherent reasoning abilities achieve significantly better results, particularly on complex temporal questions. These results demonstrate the strong generalizability of RTQA, which can effectively integrate with various LLMs in a plug-and-play manner, consistently outperforming baseline models across multiple dimensions.

\paragraph{Efficiency Analysis}
We evaluate RTQA efficiency using test questions from MultiTQ and TimelineKGQA, measuring \texttt{Avg Depth}, \texttt{Avg Branch}, and \texttt{Avg API Call}. As shown in Table~\ref{tab:efficiency}, MultiTQ questions are simpler (depth: 1.37, branch: 1.60) than those in TimelineKGQA (depth: 1.57, branch: 1.81). MultiTQ requires 3.96 API calls on average, while TimelineKGQA requires 5.38 due to more extensive answer aggregation. These results show that RTQA operates efficiently with low overhead across different question complexities.

\paragraph{Effect of Context Limits}

We analyzed the impact of context length on answer accuracy and completeness. The hyperparameter \texttt{n} controls the number of top-ranked facts retrieved as context. As shown in Table~\ref{tab:topn}, Recall@\texttt{n} increases monotonically with larger \texttt{n}. However, Hits@1 first improves and then drops, as excessive irrelevant information introduces noise and impedes LLM reasoning. Given the nearly half a million candidate facts, a sufficiently large context length is necessary. In practice, we observed that setting \texttt{n}=50 achieves the best accuracy.

\begin{table}[t]
\caption{Effect of Context Limits on Answer Quality}
\label{tab:topn}
\centering
\small
\begin{tabular}{cccc}
\toprule
 \textbf{Context Length} & \textbf{Hits@1} & \textbf{Recall@n} \\
\midrule
n=10 & 70.4\% & 53.45\% \\
n=20 & 73.8\% & 62.42\% \\
n=30 & 76.0\% & 66.44\% \\
n=40 & 76.2\% & 69.54\% \\
n=50 & 77.8\% & 71.78\% \\
n=60 & 76.0\% & 73.95\% \\
\bottomrule
\end{tabular}
\end{table}

\subsection{Case Study}
\begin{figure}[t]
    \centering
    \includegraphics[width=\linewidth]{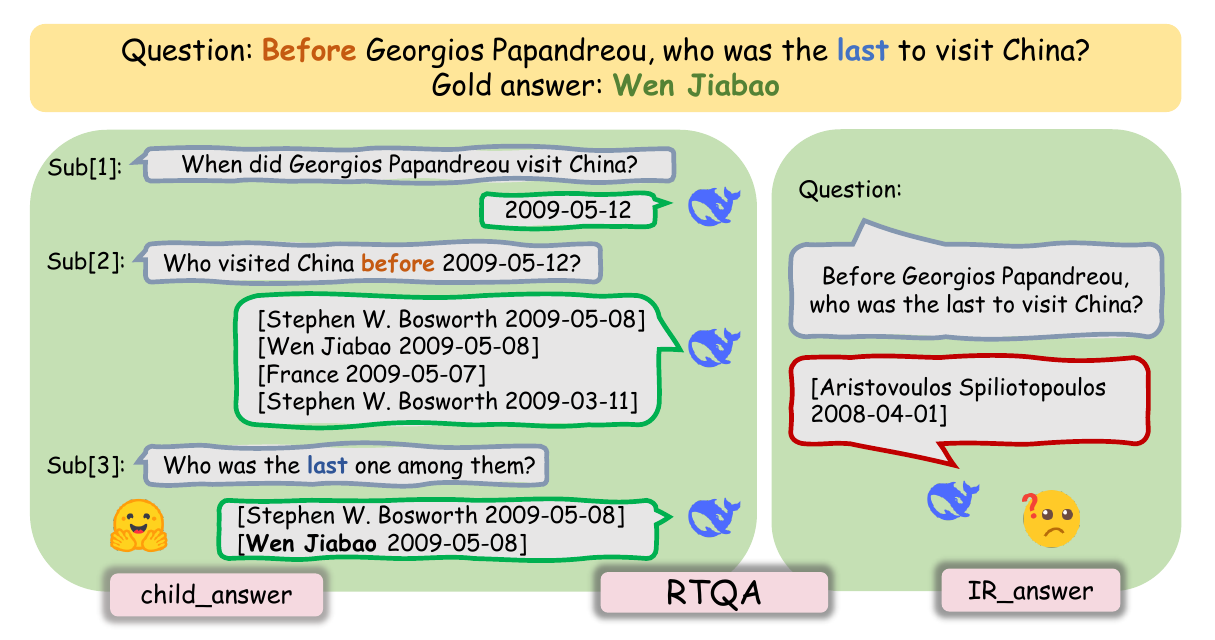}
     \vspace{-2em}
    \caption{Case study of RTQA.}
    \label{fig:case}
    \vspace{-1em}
\end{figure}

Figure~\ref{fig:case} compares two reasoning strategies of RTQA on the same question: solving via direct reasoning and solving via recursive sub-question decomposition. The comparison highlights the critical role of the question decomposition module in helping the model understand complex temporal constraints and generate reliable reasoning paths.
On the right side of Figure~\ref{fig:case}, RTQA fails to handle temporal constraints such as \textit{``before''} and \textit{``last''}, resulting in hallucinated answers (highlighted with red boxes). In contrast, the left side demonstrates how RTQA decomposes the question into three sub-questions and recursively solves them step by step, ultimately arriving at the correct answer.
\paragraph{Error Analysis}
Our error analysis highlights five key issues affecting performance: (1) \textbf{Evaluation errors}, where predictions using aliases of the gold answer are incorrectly marked as wrong despite normalization; (2) \textbf{Annotation errors}, caused by gold answers annotated as \texttt{None}, rendering evaluations invalid; (3) \textbf{Retrieval errors}, due to irrelevant or missing facts, leading to reasoning failures; (4) \textbf{Temporal reasoning failures}, where complex constraints cause LLM hallucinations; and (5) \textbf{Decomposition errors}, resulting from illogical or unexecutable sub-question formats. These issues underscore the need to improve evaluation protocols, annotation quality, retrieval precision, temporal reasoning, and question decomposition.

\section{Conclusion}
We present \textbf{RTQA}, a training-free TKGQA framework that tackles complex temporal queries by recursively decomposing questions into sub-questions and reasoning bottom-up with TKG knowledge. Its multi-path answer aggregation mitigates error propagation, ensuring robust performance. Experiments on MultiTQ and TimelineKGQA benchmarks demonstrate significant Hits@1 improvements in "\textit{Multiple}" and "\textit{Complex}" categories, outperforming state-of-the-art methods. RTQA’s plug-and-play design enhances compatibility with various LLMs, offering broad applicability. Future work will explore optimized decomposition and extensions to other knowledge graph domains, advancing efficient temporal question answering.

\section*{Limitations}
Despite the strong performance of RTQA, several limitations remain that warrant further improvement. Firstly, the effectiveness of question decomposition heavily depends on the capabilities of the underlying LLM. Smaller models may struggle to generate high-quality sub-questions, thereby constraining the performance of the recursive solving process. Secondly, RTQA relies on a robust retriever to gather relevant TKG facts. Failure to retrieve key information can significantly reduce the reasoning accuracy of the LLM. Lastly, our method is primarily tailored for complex temporal knowledge graph question answering, and its applicability to other QA domains has yet to be thoroughly validated. Future work should focus on enhancing model adaptability across different LLMs and question domains, improving retrieval performance, and extending the framework to broader QA tasks.

\section*{Ethics Statement}
In this paper, we investigate temporal knowledge graph question answering (TKGQA), focusing on complex reasoning over structured temporal data. Our method is developed and evaluated using publicly available and widely used datasets, including MultiTQ and TimelineKGQA. These datasets are constructed from open sources and do not contain any sensitive or personally identifiable information. Therefore, we believe that our work does not pose any ethical concerns.

\section*{Acknowledgements}
This work is founded by National Natural Science Foundation of China (NSFC62306276/\allowbreak NSFCU23B2055/\allowbreak NSFCU19B2027), Zhejiang Provincial Natural Science Foundation of China (No. LQ23F020017), Yongjiang Talent Introduction Programme (2022A-238-G), and Fundamental Research Funds for the Central Universities (226-2023-00138). This work was supported by Ant Group.

\bibliography{custom}

\appendix

\section{Dataset Details}
\label{A:dataset}
\subsection{\textsc{MultiTQ}}
MultiTQ is the largest known TKGQA dataset, constructed from the ICEWS05-15 dataset~\citep{garcia2018learning}, and contains 500K unique question-answer pairs. In addition, \textsc{MultiTQ} features multiple temporal granularities, including years, months, and days, with questions spanning over 3,600 days. The distribution of questions across categories is shown in Table~\ref{tab:multiTQ_stats}. 

\subsection{\textsc{TimelineKGQA}}
TimelineKGQA is an open-source automated QA pair generator for temporal knowledge graphs. Using TimelineKGQA, \cite{sun2025timelinekgqa} creates two benchmark datasets from the ICEWS Coded Event Data~\cite{DVN/28075_2015}(Time Range) and CronQuestion knowledge graph(Time Point) for demonstrating the question difficulty aligns with complexity categorization. The distribution of questions across categories is shown in Table~\ref{tab:timelinekgqa_static}.
\begin{table}[h]
\caption{Statistics of question categories in MULTITQ.}
\label{tab:multiTQ_stats}
\centering
\small
\begin{tabular}{lcccc}
\toprule
\multicolumn{2}{c}{\textbf{Category}} & \textbf{Train} & \textbf{Dev} & \textbf{Test} \\
\midrule
\multirow{3}{*}{Single} 
  & Equal & 135,890 & 18,983 & 17,311 \\
  & Before/After & 75,340 & 11,655 & 11,073 \\
  & First/Last & 72,252 & 11,097 & 10,480 \\
\cmidrule(lr){1-5}
\multirow{3}{*}{Multiple} 
  & Equal Multi & 16,893 & 3,213 & 3,207 \\
  & After First & 43,305 & 6,499 & 6,266 \\
  & Before Last & 43,107 & 6,532 & 6,247 \\
\cmidrule(lr){1-5}
& Total & 386,787 & 587,979 & 54,584 \\
\bottomrule
\end{tabular}

\end{table}

\begin{table}[h]
\caption{Statistics of question categories in TimelineKGQA.}
\label{tab:timelinekgqa_static}
\centering
\small
\setlength{\tabcolsep}{4pt}
\begin{tabular}{ccccc}
\toprule
\textbf{Source KG} & {} &\textbf{Train} & \textbf{Val} & \textbf{Test} \\
\midrule
\multirow{4}{*}{\textbf{CronQuestionKG}}  &Simple & 7,200 & 2,400 & 2,400 \\
\quad &Medium & 8,252 & 2,751 & 2,751 \\
\quad &Complex & 9,580 & 3,193 & 3,193 \\
\cmidrule(lr){2-5} 
 &Total & 25,032 & 8,344 & 8,344 \\
\bottomrule
\end{tabular}

\end{table}

\section{Case Study Details}
To illustrate how RTQA decomposes a complex temporal question and recursively solves it to obtain the correct answer, we analyze the question: "Before Georgios Papandreou, who was the last to visit China?" The process involves breaking the question into sub-questions, solving each recursively, and aggregating the results. Tables~\ref{tab:example-sub1},~\ref{tab:example-sub2},~\ref{tab:example-sub3},~\ref{tab:example-sub4} detail the reasoning process for each sub-question and the root node.

\begin{table*}[]
\caption{Reasoning process for sub-question idx 0.}
\label{tab:example-sub1}
\centering
\small
\begin{tabular}{p{0.2\textwidth} p{0.75\textwidth}}
\toprule
\textbf{Field} & \textbf{Content} \\
\midrule
\textbf{idx} & 0 \\
\textbf{question\_text} & When did Georgios Papandreou visit China? \\
\textbf{fa} & 3 \\
\textbf{question} & When did Georgios Papandreou visit China? \\
\textbf{IR\_answer} & 2009-05-12 \\
\midrule
\textbf{Historical Facts} & 
Georgios Papandreou made a visit to China on 2009-05-12. \par
China hosted a visit from Georgios Papandreou on 2009-05-12. \par
Georgios Papandreou expressed intent to meet or negotiate with China on 2009-05-12. \par
Georgios Papandreou made a visit to France on 2005-02-11. \par
Georgios Papandreou made a visit to France on 2010-03-05. \par
Georgios Papandreou made a visit to France on 2011-05-28. \par
Georgios Papandreou made a visit to France on 2011-03-19. \par
Georgios Papandreou made a visit to France on 2010-02-10. \\
\midrule
\textbf{answer} & 2009-05-12 \\
\bottomrule
\end{tabular}

\end{table*}

\begin{table*}[t]
\caption{Reasoning process for sub-question idx 1.}
\label{tab:example-sub2}
\centering
\small
\begin{tabular}{p{0.2\textwidth} p{0.75\textwidth}}
\toprule
\textbf{Field} & \textbf{Content} \\
\midrule
\textbf{idx} & 1 \\
\textbf{question\_text} & Who visited China before \#1? \\
\textbf{fa} & 3 \\
\textbf{question} & Who visited China before 2009-05-12? \\
\textbf{IR\_answer} & [Stephen W. Bosworth 2009-05-08], [Wen Jiabao 2009-05-08], [France 2009-05-07], [Stephen W. Bosworth 2009-03-11] \\
\midrule
\textbf{Historical Facts} &
South Korea hosted a visit from China on 2009-05-12. \par
Stephen W. Bosworth made a visit to China on 2009-05-13. \par
Lawrence Cannon made a visit to China on 2009-05-12. \par
China made a visit to South Korea on 2009-05-12. \par
Abdullah Gül hosted a visit from China on 2009-05-12. \par
Stephen W. Bosworth made a visit to China on 2009-05-08. \par
Kuomintang made a visit to China on 2009-05-27. \par
Lawrence Cannon made a visit to China on 2009-05-13. \par
Stephen W. Bosworth made a visit to China on 2009-05-15. \par
Wen Jiabao made a visit to China on 2009-05-09. \par
Kuomintang made a visit to China on 2009-05-26. \par
China hosted a visit from Iran on 2009-10-15. \par
Wen Jiabao made a visit to China on 2009-05-08. \par
China hosted a visit from France on 2009-05-07. \par
Ma Biao made a visit to China on 2009-07-05. \par
Georgios Papandreou made a visit to China on 2009-05-12. \par
China made a visit to Kazakhstan on 2009-06-12. \par
Abhisit Vejjajiva made a visit to China on 2009-06-15. \par
Wen Jiabao made a visit to China on 2009-05-28. \par
Wu Po-hsiung made a visit to China on 2009-05-25. \par
Eric Chu made a visit to China on 2009-05-17. \par
Barack Obama made a visit to China on 2009-10-12. \par
Abhisit Vejjajiva made a visit to China on 2009-06-25. \par
Xi Jinping made a visit to China on 2009-06-22. \par
China hosted a visit from the Russian military on 2009-07-11. \\
\midrule
\textbf{answer} & [Stephen W. Bosworth 2009-05-08], [Wen Jiabao 2009-05-08], [France 2009-05-07], [Stephen W. Bosworth 2009-03-11] \\
\bottomrule
\end{tabular}

\end{table*}

\begin{table*}[h]
\caption{Reasoning process for sub-question idx 2.}
\label{tab:example-sub3}
\centering
\small
\begin{tabular}{p{0.2\textwidth} p{0.75\textwidth}}
\toprule
\textbf{Field} & \textbf{Content} \\
\midrule
\textbf{idx} & 2 \\
\textbf{question\_text} & Who was the last one among them? \\
\textbf{fa} & 3 \\
\textbf{question} & Who was the last one among them? \\
\textbf{IR\_answer} & [Stephen W. Bosworth 2009-05-08], [Wen Jiabao 2009-05-08] \\
\midrule
\textbf{Relevant Facts} &
[Stephen W. Bosworth 2009-05-08]\par
[Wen Jiabao 2009-05-08]\par
[France 2009-05-07]\par
[Stephen W. Bosworth 2009-03-11]\\
\midrule
\textbf{answer} & [Stephen W. Bosworth 2009-05-08], [Wen Jiabao 2009-05-08] \\
\bottomrule
\end{tabular}

\end{table*}

\begin{table*}[h]
\caption{Reasoning process for the root node (original question).}
\label{tab:example-sub4}
\centering
\small
\begin{tabular}{p{0.2\textwidth} p{0.75\textwidth}}
\toprule
\textbf{Field} & \textbf{Content} \\
\midrule
\textbf{idx} & 3 \\
\textbf{question\_text} & Before Georgios Papandreou, who was the last to visit China? \\
\textbf{sons} & 0,1,2 \\
\textbf{gold\_answer} & Wen Jiabao \\
\textbf{question} & Before Georgios Papandreou, who was the last to visit China? \\
\textbf{IR\_answer} & [Aristovoulos Spiliotopoulos 2008-04-01] \\
\textbf{child\_answer} & [Stephen W. Bosworth 2009-05-08], [Wen Jiabao 2009-05-08] \\
\midrule
\textbf{Historical Facts} &
Georgios Papandreou made a visit to China on 2009-05-12. \par
China hosted a visit from Georgios Papandreou on 2009-05-12. \par
Georgios Papandreou expressed intent to meet or negotiate with China on 2009-05-12. \par
Wen Jiabao made a visit to Georgios Papandreou on 2010-10-10. \par
Georgios Papandreou hosted a visit from Wen Jiabao on 2010-10-10. \par
Georgios Papandreou made a visit to France on 2010-03-05. \par
Georgios Papandreou made a visit to France on 2005-02-11. \par
Georgios Papandreou made a visit to France on 2011-05-28. \par
Georgios Papandreou made a visit to France on 2011-03-19. \par
Georgios Papandreou made a visit to France on 2010-03-04. \par
Georgios Papandreou made a visit to France on 2010-02-10. \par
France hosted a visit from Georgios Papandreou on 2005-02-11. \par
Georgios Papandreou made a visit to France on 2010-03-07. \par
Georgios Papandreou made a visit to France on 2010-03-06. \par
Georgios Papandreou made a visit to France on 2006-11-26. \par
Georgios Papandreou made a visit to France on 2009-04-24. \par
Georgios Papandreou made a visit to France on 2011-11-01. \par
Middle East made a visit to Georgios Papandreou on 2008-07-01. \par
France hosted a visit from Georgios Papandreou on 2010-03-05. \par
France hosted a visit from Georgios Papandreou on 2010-02-10. \par
Antanas Valionis made a visit to China on 2006-04-20. \par
Georgios Papandreou made a visit to Iran on 2006-06-30. \par
Aristovoulos Spiliotopoulos made a visit to China on 2008-04-01. \par
Nicos Anastasiades made a visit to China on 2015-10-19. \par
Nicos Anastasiades made a visit to China on 2015-10-18. \\
\midrule
\textbf{answer} & [Stephen W. Bosworth 2009-05-08], [Wen Jiabao 2009-05-08] \\
\bottomrule
\end{tabular}

\end{table*}

\section{Prompts}
\subsection{Prompts for Temporal Question Decomposer}
\label{A:prompts for decom}
In the MultiTQ dataset, temporal questions are divided into \textbf{simple} and \textbf{multiple} categories based on complexity. The \textbf{simple} category includes \textit{equal}, \textit{first\_last}, and \textit{before\_after}, while \textbf{multiple} comprises \textit{equal multi}, \textit{before\_last}, and \textit{after\_first}. We designed category-specific prompts to guide the LLM in effective question decomposition. Figure~\ref{fig:prompt equal} presents the prompts for \textit{Simple}, including instructions and examples. Figure~\ref{fig:prompt before after} shows the prompt for \textit{Multiple}. Following the decomposition guidelines in~\cite{cao-etal-2023-probabilistic}, we adapted prompts for temporal scenarios, using manually crafted question-answer pairs from validation set.
\begin{figure*}[h]
\begin{tcolorbox}[
    colback=black!10, 
    colframe=black!70, 
    boxrule=1pt, 
    arc=5mm, 
    boxsep=2mm, 
    left=5mm, 
    right=5mm, 
    top=5mm, 
    bottom=4mm,
    width=\textwidth, 
    nobeforeafter, 
    colupper=black 
]
{\small\sffamily 
\textbf{Instruction:} 

Convert the following question into a JSON object where the question is the key and the value is an empty list. Do not include any explanation or extra text. Just return the JSON.
Just return the modified question in JSON format with an empty list as its value.

\vspace{0.5em}
\textbf{Here are a few examples:}\\[0.5em]
\textbf{Q:} Who visited France in 2009-05? \\
\textbf{A:} \{``Who visited France in 2009-05?'': []\} \\[0.5em]
\textbf{Q:} When did Qatar pay a visit to Barack Obama? \\
\textbf{A:} \{``When did Qatar pay a visit to Barack Obama?'': []\} \\[0.5em]
\textbf{Q:} Who applied for Iran in January 2010? \\
\textbf{A:} \{``Who applied for Iran in 2010-01?'': []\} \\[0.5em]
\textbf{Q:} Which country negotiated with Japan on 19 April 2005? \\
\textbf{A:} \{``Which country negotiated with Japan on 2002-04-19?'': []\} \\[0.5em]
\textbf{Q:} Who visited Japan in April 2012? \\
\textbf{A:} \{``Who visited Japan in 2012-04?'': []\} \\[0.5em]
\textbf{Q:} In May 2009, who signed an agreement with Iran? \\
\textbf{A:} \{``In 2009-05, who signed an agreement with Iran?'': []\} \\[0.5em]
\textbf{Q:} Who accused Iran in 2015? \\
\textbf{A:} \{``Who accused Iran in 2015?'': []\} \\[0.5em]
\textbf{Q:} On 19 March 2006, who threatened Iran? \\
\textbf{A:} \{``On 2006-03-19, who threatened Iran?'': []\} \\[0.5em]
\textbf{Q:} Who visited Guatemala on 7 July 2007? \\
\textbf{A:} \{``Who visited Guatemala on 2007-07-07?'': []\} \\[0.5em]
Remaining examples ...
\vspace{0.5em}

\textbf{Q:}  \\
\textbf{A:} 
} 
\end{tcolorbox}
\caption{Prompt example of RTQA for Temporal Question Decomposition, the category is \textbf{Simple} in MultiTQ.}
\label{fig:prompt equal}
\end{figure*}

\begin{figure*}[!h]
\begin{tcolorbox}[
    colback=black!10, 
    colframe=black!70, 
    boxrule=1pt, 
    arc=5mm, 
    boxsep=2mm, 
    left=5mm, 
    right=5mm, 
    top=5mm, 
    bottom=4mm,
    width=\textwidth, 
    nobeforeafter, 
    colupper=black 
]
{\small\sffamily 
\textbf{Instruction:} 

You are an expert specializing in dealing with problems containing the keywords "before/after". You need to read the question carefully. 

1.If the problem involves a situation like "before December 13, 2005" with a "before+ timestamp", there is no need to decompose the original problem. Just convert the question into a JSON object where the question is the key and the value is an empty list. 

2.If the problem involves the situation of a "before+ entity" like "before Japan", the original problem needs to be decomposed into sub-problems. First, generate an explicit sub-question to determine the time (e.g., "When did Iran…?"). When a sub-question is logically depends on the answer to a previous one, use placeholders (e.g., \#1) to refer to that answer. Return a valid JSON object representing the question tree. Each key is a parent question, and its value is a list of sub-questions.

\vspace{0.5em}
\textbf{Here are a few examples:}\\[1em]
\textbf{Q:} Who rejected Iran before the citizens of State Actor did? \\
\textbf{A:} \{``Who rejected Iran before the citizens of State Actor did?'': [``When did the citizens of State Actor reject Iran?'', ``Who rejected Iran before \#1?'']\} \\[0.5em]
\textbf{Q:} After Japan, who made South Korea suffer from conventional military forces? \\
\textbf{A:} \{``After Japan, who made South Korea suffer from conventional military forces?'': [``When did Japan make South Korea suffer from conventional military forces?'', ``Who make South Korea suffer from conventional military forces after \#1?'']\} \\[0.5em]
\textbf{Q:} Which country did Qatar appeal to after April 2011? \\
\textbf{A:} \{``Which country did Qatar appeal to after 2011-04?'': []\} \\[0.5em]
\textbf{Q:} Before 14 October 2015, who made Burundi suffer from conventional military forces? \\
\textbf{A:} \{``Before 2015-10-14, who made Burundi suffer from conventional military forces?'': []\} \\[0.5em]
\textbf{Q:} Who had a telephone conversation with Japan after November 2005? \\
\textbf{A:} \{``Who had a telephone conversation with Japan after 2005-11?'': []\} \\[0.5em]
\textbf{Q:} Who negotiated with Colombia before 22 December 2010? \\
\textbf{A:} \{``Who negotiated with Colombia before 2010-12-22?'': []\} \\[0.5em]
\textbf{Q:} With which country did Qatar sign formal agreements before 15 January 2008? \\
\textbf{A:} \{``With which country did Qatar sign formal agreements before 2008-01-15?'': []\} \\[0.5em]
\textbf{Q:} After November 2007, who wanted to engage in diplomatic cooperation with Timor-Leste? \\
\textbf{A:} \{``After 2007-11, who wanted to engage in diplomatic cooperation with Timor-Leste?'': []\} \\[0.5em]
\textbf{Q:} Before 24 January 2005, who wanted to establish diplomatic cooperation with the Kuomintang? \\
\textbf{A:} \{``Before 2005-01-24, who wanted to establish diplomatic cooperation with the Kuomintang?'': []\} \\[0.5em]
\textbf{Q:} Who negotiated with Bolivia after June 2007? \\
\textbf{A:} \{``Who negotiated with Bolivia after 2007-06?'': []\} \\[0.5em]
Remaining examples ...
\vspace{0.5em}

\textbf{Q:}  \\
\textbf{A:} 
} 
\end{tcolorbox}
\caption{Prompt example of RTQA for Temporal Question Decomposition, the category is \textbf{Multiple} in MultiTQ.}
\label{fig:prompt before after}
\end{figure*}

\subsection{Prompts for Recursive Solver}
\label{A:prompts for solver}
Figure~\ref{fig:prompt solver historal} shows the prompt used in the initial step of the recursive solving process. The prompt provides the large language model (LLM) with the original complex temporal question and historical facts retrieved from the temporal knowledge graph (TKG). The LLM is tasked with reasoning over these facts to either decompose the question into sub-questions or directly provide an answer if the question is simple enough. For example, for the question "Who was the president of the United States when Barack Obama became a senator?", the prompt includes historical facts such as Barack Obama’s timeline (e.g., "Barack Obama became a senator in 2005") and U.S. presidential terms (e.g., "George W. Bush was president from 2001 to 2009"), enabling the LLM to perform temporal reasoning.

Figure~\ref{fig:prompt solver relevant} depicts the prompt used in a subsequent step of the recursive solving process. The prompt supplies the LLM with the original question (or a sub-question) and relevant facts determined from the previous sub-question’s answer, asking the LLM to make the most accurate choice for the current step. This builds on the recursive decomposition by leveraging prior answers to resolve temporal dependencies. For instance, for the question "Which country was the last one among them?", if the previous sub-question "List the countries and their independence dates: [France: 1789, Germany: 1871, Japan: 1945]" yields these facts, the prompt provides this data, and the LLM returns "Japan" as the country with the latest independence date (1945).
\begin{figure*}[!h]
\begin{tcolorbox}[
    colback=black!10, 
    colframe=black!70, 
    boxrule=1pt, 
    arc=5mm, 
    boxsep=2mm, 
    left=5mm, 
    right=5mm, 
    top=5mm, 
    bottom=4mm,
    width=\textwidth, 
    nobeforeafter, 
    colupper=black 
]
{\small\sffamily 
\textbf{Instruction:} 

Based on the historical facts, please answer the given question clearly in the following format: ...So the answer is: <final concise answer>.

1.If the question asks for a specific year (e.g., "Which year", "In which year", "the exact year", etc.), then return the answer in "yyyy" format. Just return the most appropriate timestamp as the answer.

2.If the question asks for a specific month (e.g., "Which month", "In what month", "the exact month", etc.), then return the answer in "yyyy-mm" format, including the year and the month. Just return the most appropriate timestamp as the answer.

3.If the question asks for a specific date (e.g., contains keywords like "When", "What day", "the exact date", etc.), return the answer in "yyyy-mm-dd" format. Just return the most appropriate timestamp as the answer.

4.If the question asks for a set of entities (e.g., contains keywords like "who", "which country", etc.), and multiple sources in the context offer valid answers, return the union of all correct, non-duplicate entities and attached timestamp in a list format. 

\vspace{0.5em}
\textbf{Here are a few examples:}\\[1em]
\textbf{Historical facts:}
Barack Obama Reject Party Member (United Kingdom) 2008-09-23. \\
Barack Obama Reject Party Member (United Kingdom) 2008-09-23. \\
Barack Obama Make statement Party Member (United Kingdom) 2008-11-08. \\
Barack Obama Make statement Party Member (United Kingdom) 2008-11-08. \\
Barack Obama Express intent to meet or negotiate Party Member (United Kingdom) 2009-03-10. \\
Zawahiri Reject Barack Obama 2009-08-04. \\
\textbf{Question:} In which year did Barack Obama reject the party member of United Kingdom? \\
\textbf{Answer:} The rejection event occurred on 2008-09-23, so the year is 2008. So the answer is: 2008. \\[0.5em]
\textbf{Historical facts:}
Media Personnel (Somalia) Praise or endorse Cabinet / Council of Ministers / Advisors (Somalia) 2012-11-27. \\
Media Personnel (Somalia) Praise or endorse Cabinet / Council of Ministers / Advisors (Somalia) 2015-01-12. \\
Media Personnel (Somalia) Make statement African Union 2007-10-08. \\
Media Personnel (Somalia) Make statement African Union 2007-01-17. \\
Media Personnel (Somalia) Make statement African Union 2012-11-18. \\
Cabinet / Council of Ministers / Advisors (Somalia) Praise or endorse Media Personnel (Somalia) 2011-11-28. \\
Media Personnel (Somalia) Make statement African Union 2007-06-15. \\
\textbf{Question:} When did Somalia's media personnel first commend Somalia's council of ministers? \\
\textbf{Answer:} We are asked to find the first time Somalia's media personnel commended (i.e., praised or endorsed) Somalia's Cabinet / Council of Ministers / Advisors. From the historical facts: Media Personnel (Somalia) Praise or endorse Cabinet / Council of Ministers / Advisors (Somalia) on: 2012-11-27 and 2015-01-12. Among these, the earliest instance is 2012-11-27. So the answer is: 2012-11-27. \\[0.5em]
\textbf{Historical facts:}
Agence France-Presse Demand China in 2010-05-26. \\
Agence France-Presse Make an appeal or request China in 2007-01-08. \\
China Appeal for military aid Agence France-Presse in 2008-03-26. \\
France Make an appeal or request China in 2012-06-05. \\
France Demand China in 2008-06-11. \\
\textbf{Question:} Could you tell me the exact month when Agence France-Presse appealed to China? \\
\textbf{Answer:} So the answer is: 2007-01 \\[0.5em]

Remaining examples ...
\vspace{0.5em}

\textbf{Historical facts:}  \\
\textbf{Question:} \\
\textbf{Answer:}
} 
\end{tcolorbox}
\caption{Prompt example of RTQA for Recursive Solver. This prompt utilizes the triples in the TKG retrieved by the retriever as external knowledge to assist the LLM in solving the problem.}
\label{fig:prompt solver historal}
\end{figure*}

\begin{figure*}[!h]
\begin{tcolorbox}[
    colback=black!10, 
    colframe=black!70, 
    boxrule=1pt, 
    arc=5mm, 
    boxsep=2mm, 
    left=5mm, 
    right=5mm, 
    top=5mm, 
    bottom=4mm,
    width=\textwidth, 
    nobeforeafter, 
    colupper=black 
]
{\small\sffamily 
\textbf{Instruction:} 

Based on the Relevant facts, please answer the given question clearly in the following format: ...So the answer is: <final concise answer>.

Each question provides a series of relevant facts, including "entity + timestamp" pairs. You need to choose the earliest or latest entity as the answer based on the order in which the events occurred. 

\vspace{0.5em}
\textbf{Here are a few examples:}\\[1em]
\textbf{Relevant facts: } ["China 2006-01-20", "China 2006-10-30", "Vietnam 2008-04-30"]\\
\textbf{Question:} Which country was the last one among them?\\
\textbf{Answer:} The last country among the relevant facts, based on the timestamps, is Vietnam. So the answer is: Vietnam 2008-04-30.\\[0.5em]

Remaining examples ...
\vspace{0.5em}

\textbf{Relevant facts:}  \\
\textbf{Question:} \\
\textbf{Answer:}
} 
\end{tcolorbox}
\caption{Prompt example of RTQA for Recursive Solver. This prompt is mainly used to solve the problem of choosing the best first/last solution among multiple candidate answers.}
\label{fig:prompt solver relevant}
\end{figure*}

\subsection{Prompt for Answer Aggregator}
\label{A:prompts for answer aggre}
Figure~\ref{fig:prompt answer aggre} shows the instruction and examples for answer aggregation.
\begin{figure*}[!h]
\begin{tcolorbox}[
    colback=black!10, 
    colframe=black!70, 
    boxrule=1pt, 
    arc=5mm, 
    boxsep=2mm, 
    left=5mm, 
    right=5mm, 
    top=5mm, 
    bottom=4mm,
    width=\textwidth, 
    nobeforeafter, 
    colupper=black 
]
{\small\sffamily 
\textbf{Instruction:} 

 You are given a question and multiple candidate answers from sources A, B, and C.
 
 Follow these strict rules to choose the best answer: If only sources A and B are available, prefer B's answer unless it is "Unknown" or "Error", in which case choose A. If all three sources A, B, and C are available, prefer C's answer unless it is "Unknown" or "Error", then fall back to B, and if B is also invalid, fall back to A.

\vspace{0.5em}
\textbf{Here are a few examples:}\\[1em]
\textbf{Question:} When did the citizens of Africa express their intention to establish diplomatic cooperation with Vietnam?\\
\textbf{Candidate answer:}\\
source A: 2012-09-04\\
source B: 2012-09-04\\
Source C: Unknown\\
\textbf{Output:} So the answer is: 2012-09-04\\[0.5em]
\textbf{Question:} Who was the first to praise Juan Carlos I after 2006-02-22?\\
\textbf{Candidate answer:}\\
source A: Jorge Briz Abularach\\
source B: Unknown\\
Source C: House of Representatives (Uruguay)\\
\textbf{Output:} So the answer is: House of Representatives (Uruguay)\\[0.5em]
\textbf{Question:} Who rejected the Prime Minister of India after 2012-01-03?\\
\textbf{Candidate answer:}\\
source A: Sri Lanka\\
source B: China\\
\textbf{Output:} So the answer is: China\\[0.5em]
Remaining examples ...
\vspace{0.5em}

\textbf{Question:} \\
\textbf{Candidate answer:}\\
\textbf{Output:}
} 
\end{tcolorbox}
\caption{Prompt example of RTQA for Answer Aggregator.}
\label{fig:prompt answer aggre}
\end{figure*}

\end{document}